\documentclass[
twocolumn,
]{ceurart}

\usepackage{listings}
\begin{document}

\copyrightyear{2023}
\copyrightclause{Copyright for this paper by its authors.
  Permitted under Creative Commons License Attribution 4.0
  International (CC BY 4.0).}

\conference{IberLEF 2023, September 2023, Jaén, Spain}


\title{Experimenting with UD Adaptation of an Unsupervised Rule-based Approach for Sentiment Analysis of Mexican Tourist Texts}

\author[1]{Olga Kellert}[%
email=o.kellert@udc.es
]
\cormark[1]
\fnmark[1]
\address[1]{Universidade da Coruña, Grupo LyS, CITIC, Depto. de Ciencias de la Computación y Tecnologías de la Información, 
Campus de Elviña s/n, 15071 A Coruña, Spain}

\author[2] {Mahmud Uz Zaman}[%
email=mail.mahmuduzzaman@gmail.com
]
\fnmark[2]
\address[2]{University of Göttingen, Seminar für Romanische Philologie,
Humboldtallee 19, 37073 Göttingen, Germany}

\author[3] {Nicholas Hill Matlis}[%
email=nicholas.matlis@desy.de
]
\fnmark[3]
\address[3]{Center for Free-Electron Laser Science CFEL,
Deutsches Elektronen-Synchrotron DESY, Germany}

\author[4] {Carlos Gómez-Rodríguez}[%
email=carlos.gomez@udc.es
]
\fnmark[4]
\address[4]{Universidade da Coruña, Grupo LyS, CITIC, Depto. de Ciencias de la Computación y Tecnologías de la Información, 
Campus de Elviña s/n, 15071 A Coruña, Spain}

\cortext[1]{Corresponding author.}

\begin{abstract}
  This paper summarizes the results of experimenting with Universal Dependencies (UD) adaptation of an Unsupervised, Compositional and Recursive (UCR) rule-based approach for Sentiment Analysis (SA) submitted to the Shared Task at Rest-Mex 2023 (Team Olga/LyS-SALSA) (within the IberLEF 2023 conference). By using basic syntactic rules such as rules of modification and negation applied on words from sentiment dictionaries, our approach exploits some advantages of an unsupervised method for SA: (1) interpretability and explainability of SA, (2) robustness across datasets, languages and domains and (3) usability by non-experts in NLP. We compare our approach with other unsupervised approaches of SA that in contrast to our UCR rule-based approach use simple heuristic rules to deal with negation and modification. Our results show a considerable improvement over these approaches. We discuss future improvements of our results by using modality features as another shifting rule of polarity and word disambiguation techniques to identify the right sentiment words.    
\end{abstract}

\begin{keywords}
  Sentiment Analysis \sep
  Unsupervised \sep
  Rule-based \sep
  Sentiment dictionary
\end{keywords}

\maketitle

\section{Introduction}
The intensity of a sentiment expressed in words, sentences and paragraphs strongly depends on the syntactic context in which sentiment words like {\itshape nice} appear. For instance, {\itshape this hotel is very nice} expresses a stronger positive sentiment due to the intensifier {\itshape very} than just the statement {\itshape this hotel is nice} and the use of negation can reverse the polarity of the sentiment as in {\itshape this hotel is not nice}. Modification of sentiment words by intensifiers like {\itshape very} and negation are among the basic syntactic rules influencing the semantic interpretation of sentiment words.
Our main contribution in this article is a) to exploit an unsupervised (knowledge-based) model for compositional and recursive sentiment analysis (SA) driven by basic syntactic rules (Vilares et al., 2017 \cite {Vilares2017}) and to adapt it to the formalism of Universal Dependencies (UD) which is a universal framework for annotation of grammar across different human languages \cite {11234/1-5150}, b) to test our adaptation on the dataset provided by the Shared Task Rest-Mex 2023 organizers \cite{alvarez2023overview} and finally c) to compare the results of our unsupervised approach with other unsupervised methods that use heuristic rules to address modification and negation (Hutto and Gilbert, 2014 \cite {Hutto_Gilbert_2014}) as well as other teams that participated at the Shared Task Rest-Mex 2023.

The remainder of this article is structured as follows. §2 reviews related work. §3 introduces the adapted
formalism for syntactic operations. §4 presents experimental results of the Shared Task participation and compares the results with other unsupervised approaches that use heuristic rules. Finally, §5 concludes and discusses directions for future work.

\section{Related work}

The most widely used approaches to perform sentiment analysis (SA) include Machine Learning/Deep Learning based methods, lexicon-based methods and hybrid methods.

\subsection{Machine Learning/Deep Learning methods}

Supervised approaches using machine learning, deep learning and pre-trained models like BERT and RoBERTa are a common practice in SA \cite {SocPerWuChuManNgPot2013a}, \cite{cu-devlin-etal-2019-bert}, \cite{rezaeinia2017improving}, \cite {Hong2015AnalysisWD}, \cite {su15032573}, \cite {DBLP:journals/corr/abs-2110-09454}, \cite {huang-etal-2020-weakly}. In machine learning and deep learning-based approaches, the dataset is separated into training and testing datasets. A training dataset is used during the training process to learn correlations between the specific input text and the sentiment polarity. The testing dataset is then used to predict sentiment polarity on the basis of learned associations during the training period. The performance of these approaches is known to be relatively high if the learning and prediction tasks are performed on a similar dataset or corpus, as in the case of the present Shared Task of the Rest-Mex 2023 \cite{alvarez2023overview}. The downside of these approaches is that they require a huge amount of data for learning and prediction, which makes them sub-optimal for the use in real life situations where huge amounts of data are often missing or users do not have enough expertise in training and using these models. In addition, the learned associations between input and output are often obscure to be easily understood or explained.

\subsection{Lexicon-based approaches}
The lexicon-based methodology makes use of a sentiment dictionary that contains sentiment words with corresponding polarity scores in order to find out the overall opinion or polarity of a sentence or text \cite{Lexicon-BasedMethods,Hutto_Gilbert_2014,Vilares2017,VASHISHTHA2019112834}. These approaches fall into various branches depending on how they deal with syntactic rules that can shift or intensify the polarity. Non-compositional and non-recursive rule based approaches such as Vader \cite {Hutto_Gilbert_2014} use heuristic rules to account for polarity changes due to negation, modification or other syntactic processes. Other unsupervised approaches use a more general architecture to account for SA by exploiting general syntactic rules that influence sentiment polarity \cite {Vilares2017}. Lexicon-based methods are considered to be simple to use and cost effective as there is no need to train huge amounts of data. However, they must deal with syntactic rules that influence the polarity of sentiment words like negation and modification. We briefly describe here two unsupervised methods that deal with syntactic rules in a completely different way. 

\subsubsection{Non-compositional Rule Based Approach Vader}

Vader is a rule-based model for SA which is considered one of the best sentiment classifiers of its kind (Mello et al., 2022 \cite {Mello2022}). It is relatively easy to implement and does not require a large number of computational resources. Therefore, it is computationally efficient and is suitable for large-scale applications and real-time analysis. The creators of the Vader sentiment analysis system asked a number of human raters to provide sentiment scores for the sentiment words enlisted in the sentiment dictionary. The human ratings were then averaged for each word with the aim of creating a robust sentiment dictionary that is based on the opinions of many human raters. Vader was designed to handle punctuation, capitalization, emoticons, acronyms, and slang \cite {Hutto_Gilbert_2014}. For instance, capitalization of a sentiment word like  {\itshape NICE} instead of  {\itshape nice} is used as an intensifier of the word being capitalized. To account for modification or negation of sentiment words, Vader uses the distance between the sentiment word and the modifier or negation. Farther modifying words or negation words have a relatively smaller effect on the sentiment word than words in close proximity.

Vader was primarily designed for English text analysis, but it also offers a multilingual sentiment analysis via translation \cite {Hutto_Gilbert_2014}. Vader gives a probability measure between 0 and 1 (1=highest probability) for the positive, negative, neutral and compound or mixed sentiment of the input text as can be seen from the following examples in English and Spanish:

\begin{itemize}

\item \verb|English sentence| : "VADER is VERY SMART, handsome, and FUNNY!!!"
\item \verb|score output for English sentence| : 'neg': 0.0, 'neu': 0.233, 'pos': 0.767, 'compound': 0.9342
\item \verb|Spanish sentence| : ¡¡¡VADER es MUY INTELIGENTE, guapo y DIVERTIDO!!!
\item \verb|score output for Spanish sentence| : 'neg': 0.0, 'neu': 0.27, 'pos': 0.73, 'compound': 0.9387

\end{itemize}

The performance of multilingual Vader was tested on a multilingual corpus including English and Portuguese texts, which has shown better results for English than for Portuguese SA \cite {Mello2022}. 

\subsubsection{Unsupervised, Compositional and Recursive rule-based approach}

Vilares et al., 2017 \cite {Vilares2017} use the first unsupervised compositional and recursive rule-based approach for SA. Its main concept is introduced informally here. We refer the reader to the details and formalism in the paper \cite {Vilares2017}.

\begin{figure}
  \centering
  \includegraphics[width=\columnwidth]{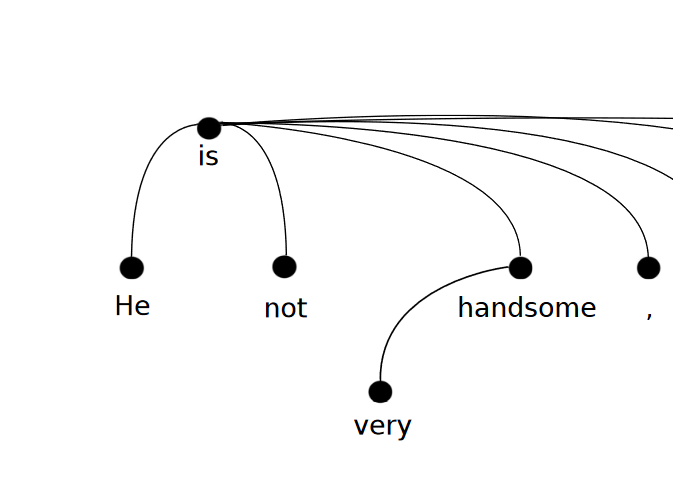}
  \caption{ A simplified version of a fragment of a dependency tree from Vilares et al., 2017}
  \label{fig:Vilares}
\end{figure}

In a nutshell, this approach exploits dependency relations between words in a sentence to deal with the scope of negation, modification and other syntactic processes that can shift, strengthen or weaken the polarity of sentiment words. In this approach, the first task is to search for sentiment words taken from a sentiment dictionary in the input text such as the sentiment word {\itshape handsome} in the sentence in Figure ~\ref{fig:Vilares} and then to traverse the dependency tree and to check whether negation words or modifiers change the polarity of sentiment words. 

The system uses a set of compositional operations to propagate changes to the semantic orientations of the nodes in the tree in a particular order. The order is basically scope driven. First, changes to polarity due to intensification apply and only after, changes to polarity due to negation apply. As a consequence, negation has scope over an intensified sentiment word as is also reflected by the hierarchical structure in Figure 1. The node of the negation word is higher than the node of the intensifier. Once all relevant operations have been executed, the processed sentiment score of the sentence is stored at the root node, which is not represented for simplicity in Figure ~\ref{fig:Vilares}.

\section{Our Experimentation with UD Adaptation of an Unsupervised Rule-based Approach}

\begin{figure*}
  \centering
  \includegraphics[width=\linewidth]{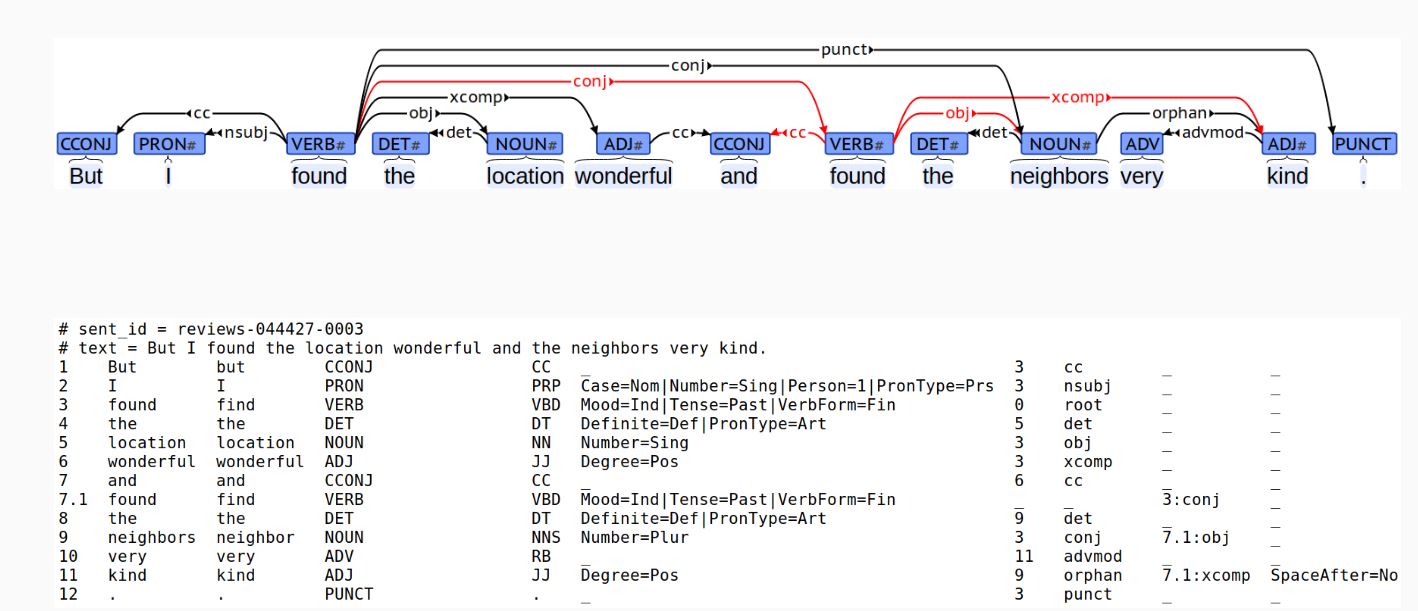}
  \caption{UD formalism. (\url{http://universaldependencies.org/eacl17tutorial/infrastructure.pdf}).}
  \label{fig:UD}
\end{figure*}

We experimented with an adaptation of an unsupervised compositional and recursive approach in SA (Vilares et al., 2017 \cite{Vilares2017} ) to the Universal Dependencies (UD) formalism \cite {11234/1-5150}, as it has since become the de facto standard for multilingual dependency parsing (the approach in \cite{Vilares2017} used Universal Treebanks instead, which was the latest precursor of UD available at that time). Figure ~\ref{fig:UD} shows a dependency structure for an English sentence and a CoNLL-U Format which represents word lines containing the annotation of a word/token with respect to various linguistic properties such as part of speech (POS), lemma, dependency relation of the word to its head, morphological features, etc. The dependency structure and linguistic properties of word/tokens as in CoNLL-U Format are an integral part of UD.

For our experimentation of an unsupervised approach to UD formalism, we used Stanza, which is a natural language toolkit based on UD-formalism that provides a basic analysis of the input text such as lemmatization, part-of-speech (POS) and dependency parsing (Peng Qi et al., 2020 (\cite {qi-etal-2020-stanza}). The dependency parser is based on UD parser from Qi et al. 2018 (\cite {qi-etal-2018-universal}. For simplicity, we do not represent all features associated with each word and focus only on those features that are relevant for SA. As we experimented with Mexican Spanish reviews from the Shared Task of Rest-Mex 2023 for our adaptation, our examples are taken from this dataset \cite{alvarez2023overview}. We did very little preprocessing of the dataset (conversion of all the uppercase letters into lowercase letters) for lemmatization and parsing reasons. Let's consider a Spanish example {\itshape No es excelente} `It is not excellent' and the associated dictionary entries with token ids, text, lemma, POS (`upos'), morphological features (`feats'), head ids and dependency relations (`deprel'): 

\begin{itemize}

\item \verb|first word| : {`id': 1, `text': `no', `lemma': `no', `upos': `ADV', `feats': `Polarity=Neg', `head': 3, `deprel': `advmod'} 
\item \verb|second word| : {`id': 2, `text': `es', `lemma': `ser', `upos': `AUX', `feats': `feats': `Mood=Ind|...', `head': 3, `deprel': `cop'} 
\item \verb|third word| : {`id': 3, `text': `excelente', `lemma': `excelente', `upos': `ADJ', `feats': `Number=Sing', `head': 0, `deprel': `root'}

\end{itemize}

Head ids and dependency relations play an important role in our approach as they provide information about the syntactic relation of words and the hierarchical structure of the sentence. Head ids contain information about parent-child relations. Take for instance, the negation word {\itshape no} and the copular word {\itshape es} in the previous example, which have {\itshape excelente} as their head. This means that the word {\itshape excelente} is the highest node and the children {\itshape no} and {\itshape es} are the lowest nodes in the structure. This head-child relation can be used to define the scope of negation. If the negation is a child of a sentiment word as its head, the polarity of the sentiment word needs to be shifted.   

In order to be able to calculate the polarity score of a sentence, we performed several steps that can be described in a nutshell as follows:

\begin{itemize}
\item \verb|Step 1| : Find sentiment words in the input text and assign polarity scores to the sentiment words
\item \verb|Step 2| : Create a dictionary of head ids and their correspondent children ids 
\item \verb|Step 3| : Identify target words that influence the sentiment word such as negation 
\item \verb|Step 4| : Calculate the polarity score for the input sentence 

\end{itemize}

Let us illustrate these steps by looking at the given Spanish example. First, we identify the sentiment word {\itshape excelente} in the input text and add new entries to the dictionary associated with this word, namely the { elementType: `count'} and the polarity score or {`elementScore': 5}. We use the dictionaries by SO-CAL for Spanish (\cite {Lexicon-BasedMethods}, \cite {Vilares2017}), in which the polarity score for sentiment words ranges from -5 (the most negative) to +5 (the most positive).

\begin{itemize}
\item \verb|Sentence| : {\itshape No es excelente} `It's not excellent'
\item \verb|Step 1| : label sentiment words
\item \verb|dictionary of the sentiment word| : {`id': 3, `text': `excelente', `lemma': `excelente', `upos': `ADJ', `feats': `Number=Sing', `head': 0, `deprel': `root', `elementType': `count', `elementScore': 5}

\end{itemize}

\begin{figure}
  \centering
  \includegraphics[width=\linewidth]{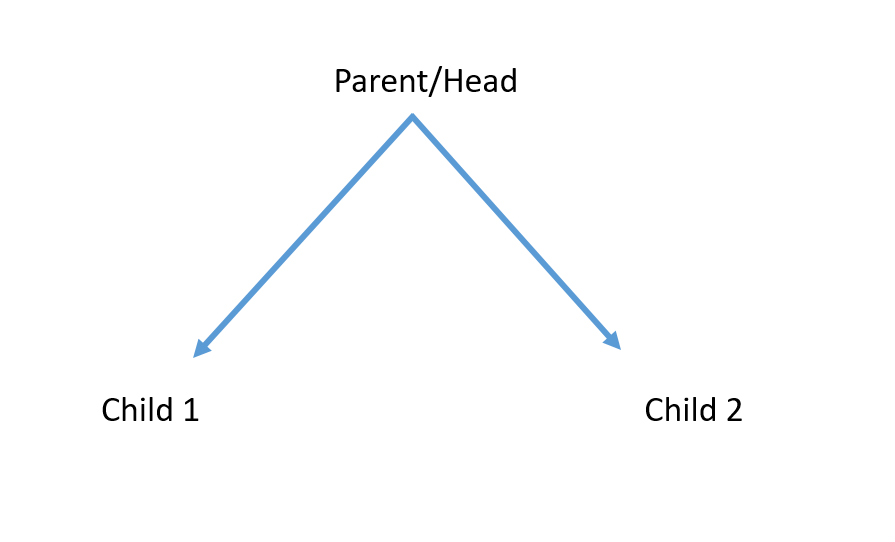}
  \caption{Example of a head-child tree branch}
  \label{fig:head-child}
\end{figure}

Step 2 consists of creating a dictionary with head ids as keys and a list of children as a key value in order to find potential polarity shifters or target words such as negation and modification. Each key-value pair of this dictionary represents a head-child tree branch as represented in Figure ~\ref{fig:head-child}.

In the UD-formalism, the head id 0 and its child represent the highest tree branch and the child of the head id 0 and its children represent the second highest branch. In our example, the second highest branch is also the lowest branch: 

\begin{itemize}
\item \verb|Sentence| : {\itshape No es excelente}
\item \verb|Step 2| : Create a dictionary with heads as keys and their correspondent children as values
\item \verb|head-child-dictionary| : {3: [1, 2], 0: [3]}

\end{itemize}

Head-child branches represent an important unit in linguistics and NLP. In linguistics, head-child branches are better known under the terms {\itshape phrases} or {\itshape maximal projections} \cite {Müller2021}, \cite {DBLP:conf/acl-cmcl/GrafM14} and they are used to describe syntactic properties or rules. In NLP, head-child branches have been used to define the domain for Machine Translation, for example \cite{DBLP:journals/corr/MengLWLJL15}. We use head-child branches for SA and more precisely to identify target words that can shift, weaken or strengthen the polarity of sentiment words. 

Step 3 consists of identifying target words that can modify the sentiment word identified in step 1. In order to achieve this goal, we loop through branches upwards and check if we can find the sentiment word, negation and/or modification in the same branch. For this, we calculate the order of branches from the lowest to the highest branch associated with a sentence. In the given sentence example {\itshape no es excelente}, the sentiment word and negation are in the same branch. 

Step 4 consists of calculating the polarity score for each branch upwards by applying the formula for the calculation of the polarity score in (1) from Vilares et al. 2017 \cite {Vilares2017}, where the variable a equals the elementScore of a sentiment word such as {\itshape excelente}, the variable b equals a value that depends on the strength of the intensifier such as {\itshape muy} taken from a list of intensifiers and negation has a score of -4 or +4 depending on the positive or negative value of a:

\begin{itemize}
\item \verb|Step 4| : Calculate the polarity score for the branch {\itshape {3: [1, 2]}}
\begin{equation}
  \ a *(1 + b)+(sign(a)*-4) = polarity score
\end{equation}
\end{itemize}

According to the formula in (1), the polarity score for the lowest branch 3:[1, 2] equals 1, if we calculate 5*(1+0)-4. As the highest branch simply expresses an identity relation between the root and the head of the previous branch, the polarity score remains the same, namely 1, and the calculation finishes with the highest branch. We take the polarity score of the highest branch to be the final result for the polarity calculation. 

\begin{figure}
  \centering
  \includegraphics[width=\linewidth]{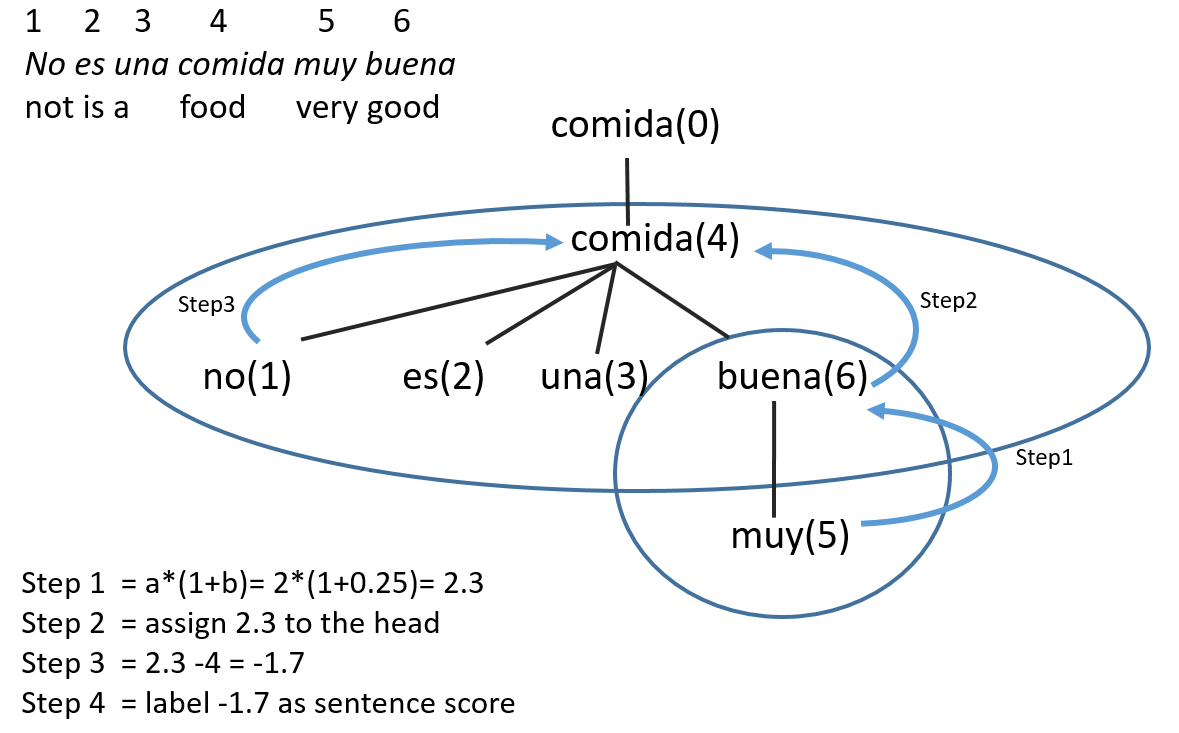}
  \caption{Example with several branches}
  \label{fig:nodes}
\end{figure}

We now discuss an example in Figure ~\ref{fig:nodes} with more branches. 

We distinguish between qualitative and quantitative modifiers. Quantitative modifiers like {\itshape big} as in {\itshape big problem} act like intensifiers as they intensify the sentiment word {\itshape problem}, similar to {\itshape very}. In our approach, quantitative modifiers can never be sentiment words and their only function is to contribute the value b in the formula (1). Qualitative adjectives like {\itshape good}, however, describe the quality of the referent expressed by the noun, they are not quantifying and they contribute the value a in the formula (1).

Turning to our example in Figure ~\ref{fig:nodes}, we first compute the intensification of the qualitative adjective {\itshape buena} `good' by the intensifier {\itshape muy} `very'. We follow the idea of assigning intensifiers like {\itshape muy} a score of 0.25 \cite {Vilares2017} . The result for this calculation is 2*(1+0.25)=2.3. We assign the score 2.3 to the nominal head {\itshape comida} `food' as the result of the nominal modification. Collecting information from the lowest branch and bringing it up to the highest branch (e.g. nominal phrase) is a common step in formal grammars such as Head-driven Phrase Structure Grammar (HPSG) \cite {Müller2021} or Minimalist Grammar (\cite {DBLP:conf/acl-cmcl/GrafM14}). As the negation is a child of the nominal head with a polarity score 2.3, the negation has scope over the nominal head. As a result, we subtract 4 from the polarity score 2.3 of the nominal head. The output of this calculation is -1.7.

The calculation finishes with the highest branch, which expresses an identity relation between the root and its child. The calculation steps are summarized as follows: 

\begin{itemize}
\item \verb|Sentence| : {\itshape No es una comida muy buena} `It's not a very good food'
\item \verb|polarity score of the lowest branch | : 2 * (1 + 0.25) = 2.3 
\item \verb|polarity score of the higher branch | : 2.3 -4 = -1.7
\item \verb|polarity score of the highest branch | : -1.7 (final polarity score) 
\end{itemize}

So far we only considered short single sentences with only one sentiment word. If the input text represents a longer sentence or several sentences with more than one sentiment word as in the case of reviews at the Shared Task Rest-Mex 2023, the polarity score needs to be calculated for the whole review text. One possibility is to calculate the mean of all sentence scores of a review (metrics a). Another option is to provide more weight to "relevant" sentences such as the last sentence of a review under the assumption that humans often say the most relevant things at the end of a review (metrics b). The third option is to consider only extreme positive or negative values (metrics c) under the assumption that humans pay more attention to extreme sentiments in a review. We illustrate the differences between the metrics by a review example and correspondent sentence scores:

\begin{itemize}
\item \verb|I did not like this hotel.| -1
\item \verb|The room was ok.| 2
\item \verb|The breakfast was mediocre.| -1
\item \verb|The service was not so great.| 1
\item \verb|In sum, not recommendable!| -4

\end{itemize}

According to the metrics b and c, the polarity for the review is -4. According to the metrics a, the review score is -0.6. For our experiments, we used the extreme value calculation (metrics c). 

\section{Experimental results}

We compare our adapted system of an unsupervised rule-based approach with another lexicon-based unsupervised approach Vader, which is extensively used in research and industry (\cite {Hutto_Gilbert_2014}, Mello et al., 2020).    
In addition, we compare our results achieved in the Shared Task (Team Olga/LyS-SALSA) with supervised approaches on the basis of the test dataset submitted to the Shared Task Rest-Mex 2023.
In both comparisons, we use accuracy as our evaluation metrics. As sentiment dictionaries, we use the dictionaries used by SO-CAL for Spanish \cite {Vilares2017} . The content of these dictionaries and their parameters are not modified or tuned. In order to calculate the sentiment score for the review, we used the extreme value metrics (metrics c).  

\subsection{Comparison with non-compositional rule based approaches}
We used the training dataset from Rest-Mex 2023 to compare our system with Vader as we do not have access to the polarity scores of the test set. The polarity scoring system used at the Shared Task is from 1 to 5, where 1 is the lowest polarity and 5 the highest. We assumed a simple mapping between our scoring system and the one used for the Shared Task (1= >-5 and <=-3, 2= >-3 and <=-1, 3= >-1 and <=1, 4= >1 and <=3, 5= >3 and <=5).
We used titles from the training dataset of Rest-Mex 2023 with positive polarity (5) and negative polarity (1) to compare the classification of our system and Vader. In order to be able to compare the polarity scores from the training dataset with Vader's scoring system that uses probability measures for polarity scores, we used three comparison metrics. In the first comparison metrics (Vader 1), we defined negative or positive scores, if the value for `pos' or `neg' in Vader's output was >0.7. In the second comparison metrics (Vader 2), the positive and negative values were defined more strictly (`pos' or `neg' >0.8) and finally in the third metrics, the probability for positive and negative scores equals 1 (Vader 3). Table~\ref{tab:fr} shows that our system (Olga/Lys-SALSA) is much better at SA than the best Vader system (see 0.27 vs. 0.62 for overall accuracy).

\begin{table*}
  \caption{Comparison between unsupervised approaches for SA based on Titles from the Training dataset Rest-Mex 2023}
  \label{tab:fr}
  \begin{tabular}{ccl}
    \toprule
    unsupervised & Accuracy\\
    \midrule
    \ Vader1 & 0.27\\
    \ Vader2 & 0.13\\
    \ Vader3 & 0.10 \\
    \ Olga/LyS-SALSA & 0.62\\
    
  \bottomrule
\end{tabular}
\end{table*}

\subsection {Comparison with supervised approaches}
For the participation at the Shared Task Rest-Mex 2023, participants were asked to classify the polarity of a test set. We used titles of the reviews as the input text for our polarity classification, if the title contained sentiment words such as "very good restaurant, but not so cheap". Otherwise, we used the full review text for polarity classification.

The best performing approaches used in the Shared Task of SA at Rest-Mex 2023 are supervised approaches which have been trained on the training dataset and which predicted polarity scores for the test dataset of the same domain or corpus as the training dataset \cite{alvarez2023overview}. It is thus not surprising to see that our results are behind the results of supervised approaches as shown in Table~\ref{tab:freq}. 

\begin{table*}
  \caption{Comparison between our system and supervised systems at Shared Task Rest-Mex 2023}
  \label{tab:freq}
  \begin{tabular}{ccl}
    \toprule
    Teams & Accuracy(Polarity)\\
    \midrule
    \ LKE-LLMAS-Team & 0.75\\
    \  Javier Alonso-Team & 0.74\\
    \ Olga/LyS-SALSA & 0.56\\
    
  \bottomrule
\end{tabular}
\end{table*}

However, it has been shown by \cite {Vilares2017} that these methods are not so robust when compared across different domains, languages and datasets. One important note regarding the comparison of our system and supervised approaches is that we do not know from the results provided to us from the organizers of the Shared Task, whether the supervised methods used both titles and reviews or just one of the two methods as the input text for the training and/or prediction task. Given that titles are very short, it is very likely that the training period was not performed on titles alone as it would probably lead to a worse performance of supervised methods. Note that our unsupervised rule-based approach does not depend much on the shortness of the text. On the contrary, in case titles contain single sentiment words like "recommendable!", no dependency parsing is required and the word score for "recommendable" directly represents the polarity score for the whole review. Finally, we would like to mention that our participation at the Shared Task was not motivated by winning the Shared Task, but by experimenting with an unsupervised rule-based approach.

\section{Discussion}

There are several issues that need to be accounted for in a lexicon-based SA system. One issue is the implementation of several syntactic rules that can shift, weaken or strengthen the polarity. One important point of this implementation is the scope of negation. We have suggested to use a head-child dependency relation to control for the scope of negation. As a result, in our approach the negation word {\itshape No} in the example {\itshape No! Es excelente!} `No! It's excellent!' is ignored, although the negation word stands in very close proximity to the sentiment word. This is because the negation and the sentiment word {\itshape excelente} do not have a head-child dependency relation. In other words, {\itshape excelente} is not the head of the negation word {\itshape No}. A similar case applies to the negation {\itshape no} in the sentence {\itshape Es excelente, no?} `It's excellent, isn't it?' Note that according to Vader, close proximity of the negation to the sentiment word counts as a trigger for shifting the sentiment polarity. As a consequence, Vader would incorrectly predict a shift in polarity for the above mentioned examples with negation. 

Scope of negation is not a trivial issue and has caused some headache to (computational) linguists, because the scope of negation does not only depend on the dependency relation, but also on semantic features of the elements negation has scope over. Take for instance, the difference between definite and indefinite nouns. Negation has usually scope over indefinite nouns, but not over definite nouns. This is because definite nouns express presuppositions and presuppositions are preserved under negation \cite{Atlas1976-ATLOTS}. Consider the relevant examples: {\itshape It's not this amazing hotel, it's the other one} vs. {\itshape It's not an amazing hotel}. Negation has scope over the adjective {\itshape amazing} in the latter, but not in the former example, because in the former example it is presupposed to be true that the hotel is amazing. We will review various cases of scope of negation in the future.

Another issue is the verb modality. Take for instance the sentiment word {\itshape great} which has a positive polarity in a sentiment dictionary. However, if the sentiment word occurs with counterfactual verbs like {\itshape would have been} as in {\itshape Everything in this hotel would have been great if it only were not so far from the center}, the verb modality weakens the polarity of the sentiment word {\itshape great}. To account for the modality, we will experiment with verbal features represented in `feats' in UD such as {feats: Mood=xx, VerbForm=yyy} in future research. We assume that capturing the verb modality of a sentence will considerably improve unsupervised SA.  

Another issue is agent-based modality. Current NLP tasks express the need for a more structured sentiment analysis that accounts for the agent-based modality (SemEval 2022, Task 10 \cite {barnes-etal-2022-semeval}). Sentiment words can be associated with different agents or opinion holders as in {\itshape According to Trip-Advisor, this hotel is great, but I don't think so.}. Quotation is another example which shows that the opinion expressed inside quotations does not need to correspond to the opinion of the reviewer as in {\itshape "this restaurant is the best" says Trip Advisor}.  We will look into modalities in more detail in the future research.

One important issue is word ambiguity. Take for instance the word {\itshape vieja} `old'. It has a negative score in our sentiment dictionary. However, if it is used in a proper name as in {\itshape Havana vieja} it does not have a negative sentiment. Depending on the domain, {\itshape old} can have a positive value as in {\itshape old tradition}. Several approaches have been dealing with Word Sense Disambiguation (WSD) including WSD for lexicon-based approaches for SA \cite{VilAloGomDocEng2013, VASHISHTHA2019112834}, most of which are exploring WordNet \cite {HUNG2016224}). Other more current approaches use neural language models for WSD (\cite {cu-giulianelli-etal-2020-analysing}). We will deal with WSD in SA in the future.   

The final issue is the calculation of the polarity score of a whole text or review that contains several sentences. We have shown that the correct prediction of sentiment scores is not only a matter of syntactic rules, but also a matter of finding the right scoring metrics that optimally represents how humans generally write reviews. As with syntactic rules that determine the syntactic context of polarity, we can assume that there are stylistic rules that determine the stylistic context of polarity. Such stylistic rules can reflect the optimal organization or structure of a text \cite {Wilson2019RelevanceT}. Marking relevant sentences by sentence or word order, key words like "in sum", capitalization, repetition, etc. can be part of these stylistic rules. Determining these rules will considerably improve unsupervised rule-based approaches and is therefore reserved for the future research.

\section{Acknowledgments}

We acknowledge the European Research Council (ERC), which has funded this research under the Horizon Europe research and innovation programme (SALSA, grant agreement No 101100615), ERDF/MICINN-AEI (SCANNER-UDC, PID2020-113230RB-C21), Xunta de Galicia (ED431C 2020/11), and Centro de Investigación de Galicia ‘‘CITIC’’, funded by Xunta de Galicia and the European Union (ERDF - Galicia 2014–2020 Program), by grant ED431G 2019/01.

\section{Appendices}


\bibliography{sample-ceur}

\appendix

\section{Online Resources}

The files used for the experiment are available on GitHub.

\begin{itemize}
\item \href{https://github.com/olga-kel/compositional-sentiment}{GitHub}

\end{itemize}

\end{document}